\documentclass[10pt]{article}
\usepackage{amsmath,amsthm,amsfonts,amssymb,latexsym,graphicx}
\usepackage{etoolbox,stmaryrd,enumitem,epigraph}
\usepackage[shortcuts]{extdash}

\allowdisplaybreaks[4]

\usepackage[authoryear]{natbib}
\bibpunct{(}{)}{;}{a}{,}{,}
\usepackage[pdfpagemode=UseNone,pdfstartview=FitH]{hyperref}

\usepackage[T2A]{fontenc}

\newtheorem{theorem}{Theorem}
\newtheorem{proposition}[theorem]{Proposition}

\theoremstyle{definition}

\theoremstyle{remark}
\newtheorem{remark}[theorem]{Remark}

\renewcommand{\d}{\mathrm{d}}
\newcommand{\dd}{\,\mathrm{d}}

\newcommand{\e}{\mathrm{e}}

\DeclareMathOperator{\E}{\mathbb{E}}

\newcommand{\Dir}{\mathrm{Dir}}

\title{Aggregation in conformal e-classification}
\author{Vladimir Vovk}

\begin{document}
\maketitle
\begin{abstract}
  Aggregating conformal predictors is a standard way of balancing
  their predictive and computational efficiency
  while retaining their validity, at least approximately.
  An important advantage of conformal e-predictors
  is that they are easier to aggregate without sacrificing their validity.
  This paper studies experimentally cross-conformal e-prediction,
  which is an existing method of aggregating conformal e-predictors,
  and its modifications that are conceptually simpler and more flexible.

  The version of this paper at \url{http://alrw.net} (Working Paper 48)
  is updated most often.
  The conference version is to appear in the COPA 2026 proceedings
  (\emph{Proceedings of Machine Learning Research}, vol.~329).
\end{abstract}

\section{Introduction}

Full conformal prediction is often computationally inefficient,
and a natural way of improving its computational efficiency
is to split the training set into two parts,
train a base predictor on one part,
and calibrate its predictions on the other part.
This approach is called inductive conformal prediction
\citep[Sect.~4.2]{Vovk/etal:2022book}
or split conformal prediction
\citep{Angelopoulos/etal:arXiv2411}.
Inductive conformal prediction typically leads
to a loss of predictive efficiency,
and a standard way to balance predictive and computational efficiency
is to aggregate several inductive conformal predictors \citep{Carlsson/etal:2014}.

A simple way of aggregating inductive conformal predictors
is cross-conformal prediction \citep[Sect.~4.4]{Vovk/etal:2022book}.
A major weakness of cross-conformal prediction
and related methods based on p-values such as jackknife+ \citep{Barber/etal:2021}
is that their provable property of validity weakens significantly
(namely, the guaranteed error probability may increase by a factor of 2,
or approximately 2)
while their typical empirical property of validity remains the same
(see, e.g., \citealt[Sects 4.4.3 and 4.8.4]{Vovk/etal:2022book}).

An important advantage of the analogue of cross-conformal prediction
based on e-values, or \emph{cross-conformal e-prediction},
is that cross-conformal e-predictors inherit the standard property of validity
of conformal e-predictors;
there is no mismatch between provable validity and typical empirical validity
\citep[Sect.~6]{Vovk:2025PR}.
This follows from the average of e-values always being an e-value,
while nothing of this kind is true for p-values
\citep{Vovk/Wang:2020}.

In this paper it will be convenient to use ``conformal prediction''
and related terms (such as ``cross-conformal prediction'')
in the generic sense, covering both methods based on p-values
(\emph{conformal p-prediction})
and methods based on e-values (\emph{conformal e-prediction}).
The term ``conformal prediction'' will also be generic in the sense of covering
full, inductive, and cross-conformal prediction.
(However, in our acronyms, such as CCP
for ``cross-conformal p-prediction'' or ``cross-conformal p-predictor'',
we will drop an extra ``P'' in order to be consistent with \citealt{Vovk/etal:2022book}.)

The main contribution of this paper is a critical discussion of cross-conformal e-prediction
and proposing its more flexible modifications.
Our experimental results will only cover simulation studies
concerning prediction of observations taking values in a finite set $\mathbf{Y}$
and generated from the standard Bayesian model with Jeffreys's prior.
(In the terminology of \citealt{Vovk/etal:2022book},
there are no objects and we are dealing with a classification problem.)
The Bayesian model will be assumed to be known,
in the spirit of the Burnaev--Wasserman programme
\citep[Sects 2.5 and 2.9.7]{Vovk/etal:2022book}.
The main topic of this paper is conformal e-prediction,
but the more standard conformal p-prediction will also be touched upon,
especially in early sections.

We start in Sect.~\ref{sec:evaluation} by defining suitable ways
of measuring predictive efficiency of conformal predictors.
This draws on and unifies the discussions in \citet[Sect.~3.1]{Vovk/etal:2022book}
and \citet[Sect.~7]{Vovk:2025PR}.
We continue in Sect.~\ref{sec:Bayes}
by formally introducing our Bayesian model,
deriving Bayes predictors for it,
and discussing ways of measuring its predictive efficiency.
Simulation results for the predictive efficiency of the Bayes predictors
are given in the following section, Sect.~\ref{sec:full},
where they are compared with results for full conformal predictors.
The definition of full conformal predictors is clear-cut only for p-predictors.
There are at least two natural definitions of full conformal e-predictors,
``ordinary'' and ``deleted'' (see Sect.~\ref{sec:full} for details).
The deleted version leads to much better results in our simulation studies,
and this is what we use in the following sections.
While conformal e-predictors are deterministic,
there are two versions of conformal p-predictors, deterministic and smoothed,
and the difference between them can be very substantial in our simulation studies
(since our datasets only involve labels, without objects, and the labels are discrete);
we report experimental results for both.

Section~\ref{sec:ICP} discusses inductive conformal prediction,
which we continue in Sect.~\ref{sec:CCP} with cross-conformal prediction.
Cross-conformal e-prediction (CCEP) has two important advantages over inductive conformal e-prediction,
the ``Jensen gap'' (which is an advantage under our criteria of predictive efficiency)
and the cross-conformal e-predictors being nearly deterministic
(having low variance);
these advantages are discussed later in the paper, in Sect.~\ref{sec:BICEP}.
There are also serious limitations of CCEP:
the fraction of the training observations that are used for calibration
in each component inductive conformal e-predictor is $1/K$,
where $K\ge2$ (the number of folds) is an integer;
therefore, first, we cannot use more than half of the training observations for calibration,
and second, we cannot have fine control of the fraction.
The limitations are inherited from CCP, but unlike CCP,
we can modify CCEP to address them.
The first limitation can be overcome by using ``inverse cross-conformal e-predictors''
(Sect.~\ref{sec:CCP}),
but a more radical approach overcoming both limitations
is to use repeated inductive conformal e-prediction (RICEP),
which involves averaging several independent inductive conformal e-predictors.
If we have an idea of a suitable proportion in which to split the training set
into proper training and calibration parts,
replacing CCEP by RICEP allows us to control the Jensen gap and the variance much better;
RICEP also allows splitting the training set in any proportion,
while for CCEP the proportions are severely limited.
This is done in Sect.~\ref{sec:BICEP}.

A disadvantage of both CCEP and RICEP is that they need the split proportion
as their parameter;
in the case of CCEP the standard parameter is the number $K$ of folds,
which corresponds to the split proportion of $(K-1):1$.
In the case of cross-validation,
a common belief is that $K=5$ or $K=10$ are reasonable choices
in wide generality.
In Sect.~\ref{sec:CCP} we will see that there are no such universal choices
in the case of CCEP,
and a natural approach is to mix over all possible proportions
with uniform weights.
This leads to ``balanced inductive conformal e-prediction'', or BICEP.
The predictive performance of RICEP and BICEP is studied in Sect.~\ref{sec:BICEP}.
Section~\ref{sec:conclusion} concludes by summarizing this paper's findings.

This paper does not contain any non-trivial theoretical results,
and its goal is modest:
to perform simulation studies of several conformal prediction algorithms
and suitable values of their parameters
(such as the number $K$ of folds in cross-conformal prediction)
and to give some recommendations.
All Jupyter notebooks required to reproduce its experimental results
can be accessed via the website
\href{http://alrw.net}{http://alrw.net}
(under Working Paper 48).

In this paper I will follow \citet{Greenland:2024}
and say ``p-surprisal'' for $-\ln p$, where $p$ is a p-value.
This terminology
(unlike ``S-value'' used by Greenland elsewhere (\citeyear{Greenland:2019}))
allows an easy extension to e-values;
namely, $\ln e$, where $e$ is an e-value, will be called ``e-surprisal''.

\section{Evaluating the predictive efficiency of conformal predictors}
\label{sec:evaluation}

Starting from the next section we will consider the simple case
where there are no objects and each observation is just a label.
But the discussion in this section will be more general.
Our goal here is to introduce suitable criteria of efficiency for conformal prediction.
In the case of conformal e-prediction,
it is the main criterion proposed in \citet{Vovk:2025PR};
we will also adapt this criterion to conformal p-prediction
obtaining a new criterion that is almost (but not quite)
a special case of the S criterion of \citet[Sect.~3.1.1]{Vovk/etal:2022book}
(as generalized in \citealt[Remark~3.15]{Vovk/etal:2022book}).

Suppose we have a training set of size $l$
and a test set with labels $y_{l+1},\dots,y_{l+k}$
(taking values in a finite set $\mathbf{Y}$).
In this paper we will evaluate the predictive efficiency
of a conformal p\-/predictor trained on the training set
by the value
\begin{equation}\label{eq:AFS}
  \frac{1}{k(\left|\mathbf{Y}\right|-1)}
  \sum_{i=l+1}^{l+k}
  \sum_{y\in\mathbf{Y}\setminus\{y_i\}}
  (-\ln p_i^y),
\end{equation}
where $p_i^y$ is the conformal p-value for the postulated label $y$
for the $i$th test observation.
Let us call it the \emph{AFS criterion} (average false p-surprisal);
it is the arithmetic average of the conformal p-surprisals for all false labels.
An attractive property of criteria of efficiency for conformal p-prediction
is being \emph{conditionally proper} \citep[Sect.~3.1.5]{Vovk/etal:2022book}.
It is reassuring that this property is satisfied by the AFS criterion;
intuitively, this means that, under this criterion,
the true conditional probability of the label given the object
is an optimal conformity measure for infinitely large training and test sets.

\begin{proposition}\label{prop:proper}
  The AFS criterion is a conditionally proper criterion of efficiency
  in the terminology of \citet[Sect.~3.1.5]{Vovk/etal:2022book}.
\end{proposition}

The proof of Proposition~\ref{prop:proper},
as well as all other proofs in this paper,
is given in Appendix~\ref{app:proofs};
it is a simple modification of the argument in \citet[Remark~3.15]{Vovk/etal:2022book}.

The use of logarithms in \eqref{eq:AFS}
has many precedents and is introduced here in analogy
with a similar expression for e-values
(cf.\ \eqref{eq:AFES} below).
For example, logarithms of p-values are summed in Fisher's method
for combining p-values \citep[Sect.~21.1]{Fisher:1925-local},
also known as Fisher's combined probability test.
Fisher's argument for this way of combining seems to have been
that p-values are akin to probabilities,
and independent probabilities combine by multiplication
\citep[the end of the first paragraph in Sect.~21.1]{Fisher:1925-local}.
Logarithms of p-values were used by \citet{Martin-Lof:1966},
and now they are ubiquitous in the algorithmic theory of randomness.
The universal p-values are defined to within an additive constant
on the logarithmic scale,
and adding logarithms of p-values features prominently
in many theorems of the algorithmic theory of randomness.
In statistics, logarithms of p-values have been widely used by
\citet{Greenland:2019,Greenland:2024} and co-authors.
S-values (i.e., p-surprisals) are believed to be additive,
at least in the independent case \citep[Sect.~4.4]{Greenland:2019}.

The analogue for e-values of the AFS criterion \eqref{eq:AFS}
is the \emph{AFES criterion}
\begin{equation}\label{eq:AFES}
  \frac{1}{k(\left|\mathbf{Y}\right|-1)}
  \sum_{i=l+1}^{l+k}
  \sum_{y\in\mathbf{Y}\setminus\{y_i\}}
  \ln e_i^y,
\end{equation}
where $e_i^y$ is the conformal e-value for the postulated label $y$
for the $i$th test observation.
It was proposed in \citet{Vovk:2025PR}
together with its simpler but less natural counterpart
\begin{equation}\label{eq:AFES-16}
  \frac{1}{k\left|\mathbf{Y}\right|}
  \sum_{i=l+1}^{l+k}
  \sum_{y\in\mathbf{Y}}
  \ln e_i^y.
\end{equation}
It is shown in \citet[Proposition~15]{Vovk:2025PR}
that in an idealized setting
with an infinite training set and $k\to\infty$
the optimal nonconformity e-measure under the AFES criterion \eqref{eq:AFES}
is very natural,
namely, the nonconformity score of an observation
is proportional to the odds against its label
(observed for a training observation or postulated for a test observation)
conditional on its object.
That proposition can be interpreted as the AFES criterion
satisfying an analogue for e-values of being conditionally proper.
And it is shown in \citet[Remark~16]{Vovk:2025PR}
that in the same idealized setting
the optimal nonconformity e-measure
under the modified AFES criterion \eqref{eq:AFES-16}
is even simpler,
namely, the nonconformity score of an observation
is proportional to the inverse of the conditional probability of its label
given its object.

The justification for the use of logarithms in \eqref{eq:AFES}
is that logarithms of e-values are additive,
as discussed by \citet[Sect.~2.2.1]{Shafer:2021}
with a reference to Kelly.
For further discussions, see \citet[Sect.~3.1]{Ramdas/etal:2023}
and \citet[Sect.~2.1]{Grunwald/etal:2024-local}.

Both criteria, \eqref{eq:AFS} and \eqref{eq:AFES},
will be used in this paper in their limiting form as $k\to\infty$.

\section{Our Bayesian model and Bayes prediction}
\label{sec:Bayes}

As mentioned earlier,
in this paper we are interested in the case of classification with no objects;
the label space is assumed to be $\mathbf{Y}=\{1,\dots,Y\}$,
and therefore, the observation space is also $\{1,\dots,Y\}$.
Our postulated model for the data-generating mechanism is Bayesian:
the parameter is $\theta\in\Theta$, the parameter space
\[
  \Theta
  :=
  \left\{
    \theta\in[0,1]^Y:\sum_{y=1}^Y\theta_y=1
  \right\}
\]
is the standard simplex,
$P_{\theta}$ is the probability measure on $\mathbf{Y}$
defined by $P_{\theta}(\{y\}):=\theta_y$, $y=1,\dots,Y$,
and $\theta\sim\Dir(\alpha,\dots,\alpha)$.
Here $\Dir(\alpha,\dots,\alpha)$ is the Dirichlet distribution
with all parameters equal to $\alpha\in(0,\infty)$,
which we will abbreviate to $\Dir_{\alpha}$.
Important special cases are $\alpha=1$ (the uniform distribution)
and $\alpha=0.5$ (Jeffreys's prior).

If the training set has size $n$ and there are $n_y$ $y$s in it
for all $y\in\mathbf{Y}$,
the probability prediction $P$ for the test observation
is
\begin{multline}\label{eq:P}
  P(y)
  =
  \frac
  {\int_{\Theta}
    \theta_y^{n_y+1}\prod_{y'\ne y}\theta_{y'}^{n_{y'}}
    \Dir_{\alpha}(\d\theta)}
  {\int_{\Theta}\prod_{y'}\theta_{y'}^{n_{y'}}\Dir_{\alpha}(\d\theta)}
  =
  \frac
  {\int_{\Theta}
    \theta_y^{n_y+\alpha}\prod_{y'\ne y}\theta_{y'}^{n_{y'}+\alpha-1}
    \dd\theta}
  {\int_{\Theta}\prod_{y'}\theta_{y'}^{n_{y'}+\alpha-1}\dd\theta}\\
  =
  \frac
  {B(n_1+\alpha,\dots,n_{y-1}+\alpha,n_y+\alpha+1,n_{y+1}+\alpha,\dots,n_Y+\alpha)}
  {B(n_1+\alpha,\dots,n_Y+\alpha)}\\
  =
  \frac
  {\Gamma(n_y+\alpha+1)/\Gamma(n+Y\alpha+1)}
  {\Gamma(n_y+\alpha)/\Gamma(n+Y\alpha)}
  =
  \frac{n_y+\alpha}{n+Y\alpha},
\end{multline}
where we drop curly braces in expressions such as $P(\{y\})$ and use $\sum_{y'}n_{y'}=n$.
This is a generalization of Laplace's rule of succession.

Let us now restate this Bayes prediction rule
in terms of p-values and e-values,
to facilitate comparison with conformal prediction.
Suppose we have a training set of size $l$.
Let $n_y$, $y\in\mathbf{Y}$, be the number of $y$s in this training set,
as before.
The optimal smoothed p-variable,
in the sense of the Bayesian analogue of the AFS criterion \eqref{eq:AFS},
that is valid under our Bayesian model is
\begin{equation}\label{eq:p}
  p_{y}
  =
  A_{y}+\tau B_{y},
  \enspace\text{where}\enspace
  A_y := \sum_{y':P(y')<P(y)} P(y'),
  \enspace 
  B_y := \sum_{y':P(y')=P(y)} P(y'),
\end{equation}
and $\tau$ is uniformly distributed on $[0,1]$;
this is the content of the following proposition.

\begin{proposition}\label{prop:Bayes-p-optimality}
  The $p_y$ defined by \eqref{eq:p} comprise
  a p-variable valid under our Bayesian model
  and attaining
  \begin{equation}\label{eq:Bayes-p-optimality}
    \E
    \sum_{y'\in\mathbf{Y}\setminus\{y\}}
    (-\ln p_{y'})
    \to
    \max
  \end{equation}
  (cf.~\eqref{eq:AFS}) in this class of p-variables,
  where the expectation is for $y_1,\dots,y_l,y$ generated
  from our Bayesian model.
\end{proposition}

Let us define the \emph{Bayesian p-predictor}
(denoted ``p-Bayes'' in figures below)
as the predictor of the same type as a conformal predictor
but outputting the optimal Bayesian p-values \eqref{eq:p}
in place of conformal p-values.
While \citet[Theorem~3.1 and Remark~3.15]{Vovk/etal:2022book} are about an idealized setting
with infinite training and test sets and IID observations,
Proposition~\ref{prop:Bayes-p-optimality} is about a finite training set.

The expected p-surprisal for \eqref{eq:p} is
\begin{equation}\label{eq:expected-ps}
  \E_{\tau}
  \left(
    -\ln p_{y}
  \right)
  =
  F(A_{y},B_{y}),
\end{equation}
where $F$ is defined as the definite integral
\begin{align}
  F(A,B)
  &:=
  -\int_0^1
  \ln(A+B\tau)
  \dd\tau
  =
  -\left[
    \left(
      \frac{A}{B}+\tau
    \right)
    \ln(A+B\tau)
    -
    \tau
  \right]_0^1\notag\\
  &=
  \frac{A}{B}
  \ln A
  +
  1
  -
  \frac{A+B}{B}
  \ln(A+B)
  \label{eq:F}
\end{align}
for $A\ge0$ and $B>0$.
In computer code
the case $A=0$ should be considered separately,
and the last expression should be replaced by $1-\ln B$.
The value of the AFS criterion \eqref{eq:AFS}
for a given training set and an infinite test set is, a.s.,
\begin{equation}\label{eq:overall-p}
  \sum_{y\in\mathbf{Y}}
  \theta_y
  \frac{1}{Y-1}
  \sum_{y'\ne y}
  F(A_{y'},B_{y'})
  =
  \frac{1}{Y-1}
  \sum_{y\in\mathbf{Y}}
  (1-\theta_y)
  F(A_{y},B_{y}),
\end{equation}
where $\theta=(\theta_y)$ is the realized parameter value
(generated from the Dirichlet distribution).
In our computer code, we average it over many simulations
of $\theta$ and the training set.

In analogy with \citet[Proposition~15]{Vovk:2025PR},
the optimal Bayesian e-value (valid under our Bayesian assumption)
for a test observation $y$ is
\begin{equation}\label{eq:e}
  e_y
  :=
  \frac{1}{Y-1}
  \frac
  {1-\frac{n_y+\alpha}{l+Y\alpha}}
  {\frac{n_y+\alpha}{l+Y\alpha}}
  =
  \frac{1}{Y-1}
  \left(
    \frac{l+Y\alpha}{n_y+\alpha} - 1
  \right).
\end{equation}
This is spelled out in the following proposition,
which extends \citet[Proposition~15]{Vovk:2025PR} from conformal to Bayesian e-prediction.

\begin{proposition}\label{prop:Bayes-e-optimality}
  The e-values $e_y$ defined by \eqref{eq:e} provide
  the only solution to the optimization problem
  \begin{equation}\label{eq:Bayes-e-optimality}
    \E
    \sum_{y'\in\mathbf{Y}\setminus\{y\}}
    \ln e_{y'}
    \to
    \max
  \end{equation}
  (cf.~\eqref{eq:AFES}),
  where the expectation is for $y_1,\dots,y_l,y$ generated
  from our Bayesian model
  and $e$ ranges over the e-variables w.r.t.\ this model.
\end{proposition}

Similarly to the case of p-values,
we define the \emph{Bayesian e-predictor}
(denoted ``e-Bayes'' in figures)
as the predictor of the same type as a conformal e-predictor
but outputting the optimal Bayesian e-values \eqref{eq:e}
in place of conformal e-values.
Similarly to Proposition~\ref{prop:Bayes-p-optimality},
Proposition~\ref{prop:Bayes-e-optimality} modifies its conformal counterpart,
Proposition~15 in \citet{Vovk:2025PR},
making it more realistic in that the training set becomes finite,
but on the negative side we have to make an extra Bayesian assumption.

Now the value of the AFES criterion \eqref{eq:AFES},
i.e., the expectation of the average conformal e-surprisal for false labels,
for a given parameter $(\theta_y)$, given training set,
and an infinite test set, is
\begin{equation}\label{eq:overall-e}
  \sum_{y\in\mathbf{Y}}
  \theta_y
  \frac{1}{Y-1}
  \sum_{y'\ne y}
  \ln e_{y'}
  =
  \frac{1}{Y-1}
  \sum_{y\in\mathbf{Y}}
  (1-\theta_y)
  \ln e_{y}
\end{equation}
a.s., where $e_y$ is defined by \eqref{eq:e}.

A simpler modification of the e-values \eqref{eq:e} is
\begin{equation}\label{eq:e-16}
  e_y
  :=
  \frac{1}{Y}
  \frac
  {1}
  {\left(\frac{n_y+\alpha}{l+Y\alpha}\right)}
  =
  \frac{1}{Y}
  \frac{l+Y\alpha}{n_y+\alpha}.
\end{equation}
It is also valid under our Bayesian model,
but is optimal in a different sense.

\begin{proposition}\label{prop:Bayes-e-optimality-16}
  The $e_y$ defined by \eqref{eq:e-16} provide the only solution
  to the optimization problem
  \[
    \E
    \sum_{y'\in\mathbf{Y}}
    \ln e_{y'}
    \to
    \max
  \]
  (cf.~\eqref{eq:AFES-16}).
\end{proposition}

The \emph{suboptimal Bayesian e-predictor}
(``suboptimal e-Bayes'' in figures)
is defined as the Bayesian e-predictor
except that it outputs \eqref{eq:e-16} instead of \eqref{eq:e}.
Of course, it is suboptimal only because of our decision
to use \eqref{eq:AFES} as our main criterion of efficiency;
it would have been optimal had we used \eqref{eq:AFES-16}.
For a given parameter $(\theta_y)$, a given training set,
and an infinite test set
the value of the modified AFES criterion~\eqref{eq:AFES-16}
is, a.s.,
\begin{equation}\label{eq:overall-e-16}
  \sum_{y\in\mathbf{Y}}
  \theta_y
  \frac{1}{Y}
  \sum_{y'\in\mathbf{Y}}
  \ln e_{y'}
  =
  \frac{1}{Y}
  \sum_{y\in\mathbf{Y}}
  \ln e_{y}.
\end{equation}
The suboptimal Bayesian e-predictor consistently achieves
better (i.e., larger) values of \eqref{eq:overall-e-16}
than the Bayesian e-predictor does
(but this is not shown in the figures below).

\section{Full conformal prediction}
\label{sec:full}

The main disadvantage of the Bayesian p-values and e-values
derived in the previous section
is that they are only valid under the postulated Bayesian model.
In this sense this Bayesian model serves as a hard model
in the terminology of \citet[Sect.~2.5.1]{Vovk/etal:2022book}.
In this and following sections we will use the Bayesian model
only for defining nonconformity measures for conformal prediction,
and the resulting p-values and e-values will be valid under exchangeability.
Only their predictive efficiency will depend on the postulated Bayesian model,
and so the latter becomes our soft model.
As mentioned earlier,
we do not claim any formal properties of predictive efficiency for our methods
and study it experimentally.

We start by deriving the full conformal p-predictor (CP).
As before, $n_y$, $y\in\mathbf{Y}$, is the number of $y$s in the training set of size $l$.
As conformity score of an observation $y$ we use an estimate of the conditional probability of $y$
or, equivalently, the number of $y$s in the comparison bag
(the resulting conformal p-value does not depend on the details of
how exactly we define the comparison bag;
cf.\ the discussion of ordinary, studentized, and deleted e-values below).
If the true test observation is $y$,
for each false test observation $y'\ne y$ the conformal p-value
is
\begin{multline}\label{eq:CP}
  p_{y'}
  =
  A_{y'}+\tau B_{y'},
  \text{ where}\\
    A_{y'}
    :=
    \frac
      {\sum_{y''\in\mathbf{Y}\setminus\{y'\}:n_{y''}<n_{y'}+1} n_{y''}}
      {l+1},
    \enspace
    B_{y'}
    :=
    \frac
    {
      \sum_{y'':n_{y''}=n_{y'}+1} n_{y''}
      +
      n_{y'}+1
    }
    {l+1},
\end{multline}
and $\tau$ is, as before, uniformly distributed on $[0,1]$.
The expected p-surprisal is then given by~\eqref{eq:expected-ps}
and~\eqref{eq:F}.
The value of the AFS criterion for a given training set and an infinite test set
is given by the right-hand side of~\eqref{eq:overall-p}.

The situation with e-values is more complicated;
the e-values can be ordinary, studentized, and deleted
(see, e.g., \citealt[Sect.~7.3.1]{Vovk/etal:2022book}),
and these versions are all different (unlike in the case of p-values).
Let $\sigma\in[0,1]$ be a parameter indicating how ordinary our conformal e-predictor is;
its precise meaning will be explained
after introducing nonconformity scores~\eqref{eq:score-typical} and \eqref{eq:score-test}.
Each of the $n_y$ observations $y$ in the training set has nonconformity score
\begin{equation}\label{eq:score-typical}
  \frac
  {1-\frac{n_y-1+\sigma+\alpha}{l+\sigma+Y\alpha}}
  {\frac{n_y-1+\sigma+\alpha}{l+\sigma+Y\alpha}}
  =
  \frac{l+\sigma+Y\alpha}{n_y-1+\sigma+\alpha} - 1
\end{equation}
in the augmented training set
(the training set augmented by the postulated test observation)
unless $y$ coincides with the postulated test observation;
if it does, the nonconformity score is
\begin{equation}\label{eq:score-test}
  \frac
  {1-\frac{n_y+\sigma+\alpha}{l+\sigma+Y\alpha}}
  {\frac{n_y+\sigma+\alpha}{l+\sigma+Y\alpha}}
  =
  \frac{l+\sigma+Y\alpha}{n_y+\sigma+\alpha} - 1;
\end{equation}
the nonconformity score of a postulated test observation $y$
is also given by \eqref{eq:score-test}.

The nonconformity scores~\eqref{eq:score-typical} and \eqref{eq:score-test}
both use the Bayesian probabilities \eqref{eq:P}.
The case $\sigma=0$ is the main one used in this paper;
it is the \emph{deleted} version of the definition.
The Bayesian training set (serving as our comparison bag)
is the augmented training set minus the observation
for which we are computing its nonconformity score.
The size of the Bayesian training set is $l$,
as in both~\eqref{eq:score-typical} and \eqref{eq:score-test}.
Both~\eqref{eq:score-typical} and \eqref{eq:score-test}
are the estimated odds against $y$.
In \eqref{eq:score-typical}, we use $n_y-1$ as the number of $y$s
since we are excluding one of the observations $y$.
And \eqref{eq:score-test} covers two cases:
if we are computing the nonconformity score of the test observation,
there is no need to decrease $n_y$ since $n_y$ only includes training observations;
and if we are computing the nonconformity score of a training observation,
we first exclude it but then we include the test observation $y$ in our count of $y$s.

For $\sigma=1$ we will have the \emph{ordinary} version,
in which an observation $y$ is included in the Bayesian training set
(our comparison bag)
when computing its nonconformity score.
Its inclusion is reflected in adding $\sigma=1$ to the numerators and denominators
of all fractions in~\eqref{eq:score-typical} and~\eqref{eq:score-test}.
The case $\sigma=0.5$ resembles the studentized version
in that it is intermediate between deleted and ordinary
(although it is between deleted and ordinary in a different sense
from the usual studentized residuals;
see \citealt[Sect.~4.2.2]{Montgomery/etal:2012}).

The nonconformity scores~\eqref{eq:score-typical} and~\eqref{eq:score-test}
give us this formula for the conformal e-value for a postulated test observation $y'$:
\begin{equation}\label{eq:e-CP}
  e_{y'}
  =
  \frac{\frac{l+\sigma+Y\alpha}{n_{y'}+\sigma+\alpha} - 1}
  {
    \frac{1}{l+1}
    \left(
      \sum_{y\ne y'}
      n_y
      \left(
        \frac{l+\sigma+Y\alpha}{n_y-1+\sigma+\alpha} - 1
      \right)
      +
      (n_{y'}+1)
      \left(
        \frac{l+\sigma+Y\alpha}{n_{y'}+\sigma+\alpha} - 1
      \right)
    \right)
  };
\end{equation}
the e-value is obtained by normalizing the nonconformity score of the test observation
by dividing it by the average nonconformity score in the augmented training set.
This defines the (full) \emph{conformal e-predictor} (CEP).

In computer code,
we should be careful with the possibility of division by zero in \eqref{eq:e-CP},
since $n_y-1+\sigma+\alpha=0$ is possible (but only when $n_y=0$).
If $n_y=0$, the fraction with $n_y-1+\sigma+\alpha=0$ in the denominator
should be set to, say, 0 (the exact value does not matter
since the fraction is multiplied by $n_y=0$).

In our code we first compute
\begin{equation}\label{eq:S}
  S
  =
  \sum_{y\in\mathbf{Y}}
  n_y
  \left(
    \frac{l+\sigma+Y\alpha}{n_y-1+\sigma+\alpha} - 1
  \right),
\end{equation}
after which we can replace \eqref{eq:e-CP} by the quicker expression
\begin{equation}\label{eq:e-CP-quick}
  e_{y'}
  =
  \frac{
    (l+1)
    \left(
      \frac{l+\sigma+Y\alpha}{n_{y'}+\sigma+\alpha} - 1
    \right)}
  {
    S
    -
    n_{y'}
    \left(
      \frac{l+\sigma+Y\alpha}{n_{y'}-1+\sigma+\alpha} - 1
    \right)
    +
    (n_{y'}+1)
    \left(
      \frac{l+\sigma+Y\alpha}{n_{y'}+\sigma+\alpha} - 1
    \right)}.
\end{equation}

\begin{figure}[bt]
  \begin{center}
    \includegraphics[width=0.48\textwidth]{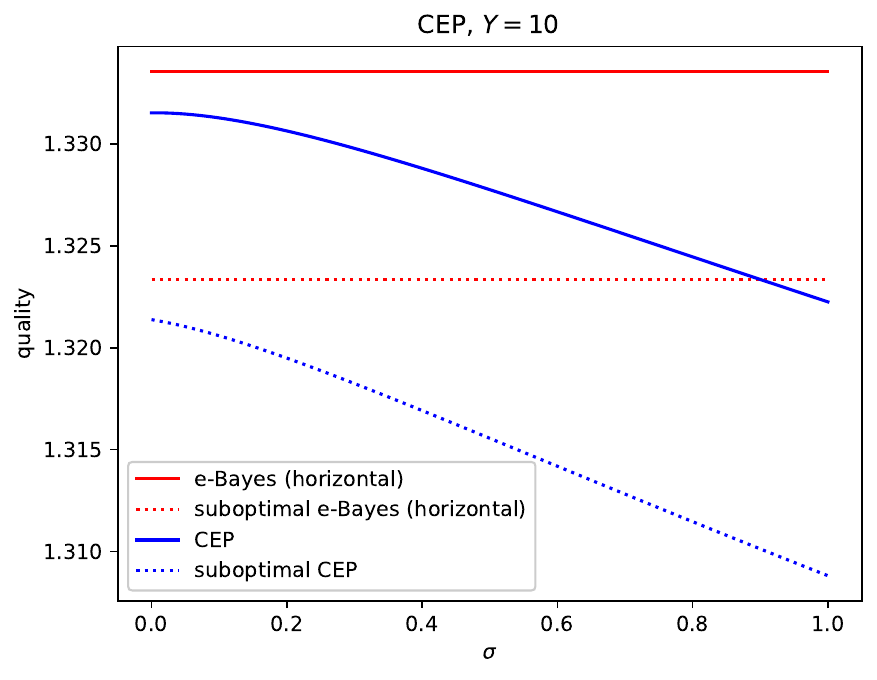}
    \includegraphics[width=0.48\textwidth]{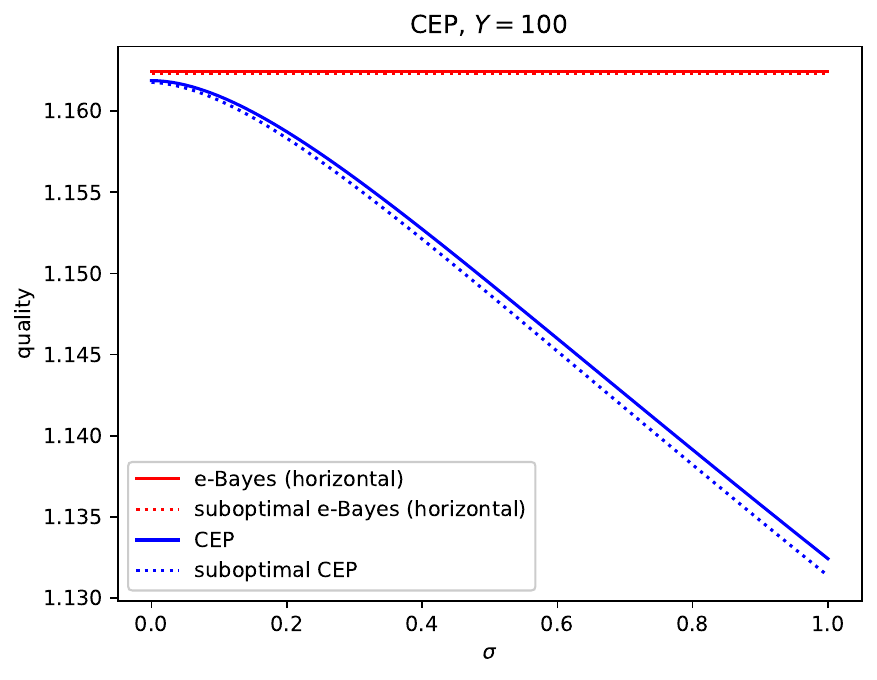}
  \end{center}
  \caption{Results for CEP with different degrees of ordinariness $\sigma\in[0,1]$;
    $l=12{,}000$, $Y=10$ (left), and $Y=100$ (right).
    The graphs show the average value of the AFES criterion over 10,000 iterations
    as function of $\sigma$.
    This is for $\alpha=0.5$ (i.e., Jeffreys's prior).}
  \label{fig:CEP}
\end{figure}

The value of the AFES criterion is given by \eqref{eq:overall-e},
and we evaluate its average over 10,000 independent draws of $\theta$ from Jeffreys's prior
and then of the training set
for various $\sigma$ in Figure~\ref{fig:CEP} (the solid graphs).
The results for $\sigma=0$ are much better than those for $\sigma=1$,
and the results for the quasi-studentized case $\sigma=0.5$ are intermediate.
In our experiments we choose $l=12{,}000$ as the size of our training set
for two reasons.
First, this choice leads to e-values comparable with Jeffreys's $\sqrt{10}$ and $10$
(on the log scale, $\ln\sqrt{10}\approx1.15$ and $\ln10\approx2.30$)
and to p-values comparable with Fisher's $5\%$ and $1\%$
(on the log scale, $\ln(1/0.05)\approx3.00$ and $\ln(1/0.01)\approx4.61$);
see \citet[Sect.~2]{Vovk/Wang:2023} for a discussion of these conventional thresholds.
Second, $12{,}000$ is convenient in the context of CCP and CCEP
since it has many small divisors $K$
(which can be conveniently used as the numbers of folds).

The nonconformity scores \eqref{eq:score-typical}--\eqref{eq:score-test}
are the counterpart of \eqref{eq:e} in full conformal e\-/prediction
(motivated by \citealt[Proposition 15]{Vovk:2025PR}).
Now let us discuss the counterpart of \eqref{eq:e-16}
(motivated by \citealt[Remark 16]{Vovk:2025PR}),
where we replace the likelihood ratio against the observation
by its inverse probability.
It involves a slight modification,
which we call \emph{suboptimal CEP}:
now we should drop all entries of ``${}-1$''
outside the expressions ``$n_y-1$'' and ``$n_{y'}-1$''
in \eqref{eq:score-typical}, \eqref{eq:score-test}, \eqref{eq:e-CP},
\eqref{eq:S}, and \eqref{eq:e-CP-quick},
and we should also drop the subtrahends in the numerators
of the left-hand sides of \eqref{eq:score-typical}, \eqref{eq:score-test}.
The quality of prediction as measured
by the AFES criterion \eqref{eq:overall-e} then suffers,
since this nonconformity measure is optimized for \eqref{eq:overall-e-16}.

The pictures in Figure~\ref{fig:CEP} cover
both \eqref{eq:e} and \eqref{eq:e-16} and their full conformal analogues,
with the suboptimal versions shown as dotted graphs.
The vertical axis is labelled ``quality'',
which refers to the AFES criterion
(or the AFS criterion in the case of p-values in some of the figures given below).
The case $\sigma=0$ (deleted CEP) remains the best
for the suboptimal versions.
While Figure~\ref{fig:CEP} only covers the cases $Y=10$ and $Y=100$,
the blue graphs also look monotonically decreasing (albeit almost horizontal)
in the case of binary classification $Y=2$,
which we often include in our experimental studies later in this paper.

From now on we always set $\sigma:=0$ in CEP and suboptimal CEP.
While our justification of this choice is empirical,
it would be interesting to have a theoretical explanation.

\section{Inductive conformal prediction}
\label{sec:ICP}

In inductive conformal prediction,
we randomly split the training set of size $l$ into two parts,
the \emph{proper training set} of size $m\in\{1,\dots,l-1\}$
and the \emph{calibration set} of size $m':=l-m$.
Let $n_y$ be the number of $y$s, $y\in\mathbf{Y}$,
in the proper training set
and $n'_y$ be the number of $y$s in the calibration set.

\subsection{ICP}

Let us derive the inductive conformal p-predictor (ICP)
corresponding to the AFS criterion of efficiency \eqref{eq:AFS}.
In the calibration set,
$n'_y$ elements have conformity score $\frac{n_y+\alpha}{m+Y\alpha}$ or, equivalently, $n_y$,
for each $y\in\mathbf{Y}$.
Suppose the true test observation is $y$.
For each false test observation $y'$, the corresponding p-value is
\begin{equation}\label{eq:A-B}
  p_{y'}
  =
  A_{y'}+\tau B_{y'}
  :=
  \frac{
    \sum_{y'':n_{y''}<n_{y'}} n'_{y''}
    +\tau
    \left(
      \sum_{y'':n_{y''}=n_{y'}} n'_{y''}+1
    \right)}
  {m'+1}
\end{equation}
(cf.\ \eqref{eq:CP}).
The expected p-surprisal is then \eqref{eq:expected-ps} with $y$ replaced by $y'$
and with $A_{y'}$ and $B_{y'}$ defined by \eqref{eq:A-B}.
The value of the AFS criterion for given proper training set, calibration set, and $\theta$
is still~\eqref{eq:overall-p}.

\begin{figure}[bt]
  \begin{center}
    \includegraphics[width=0.48\textwidth]{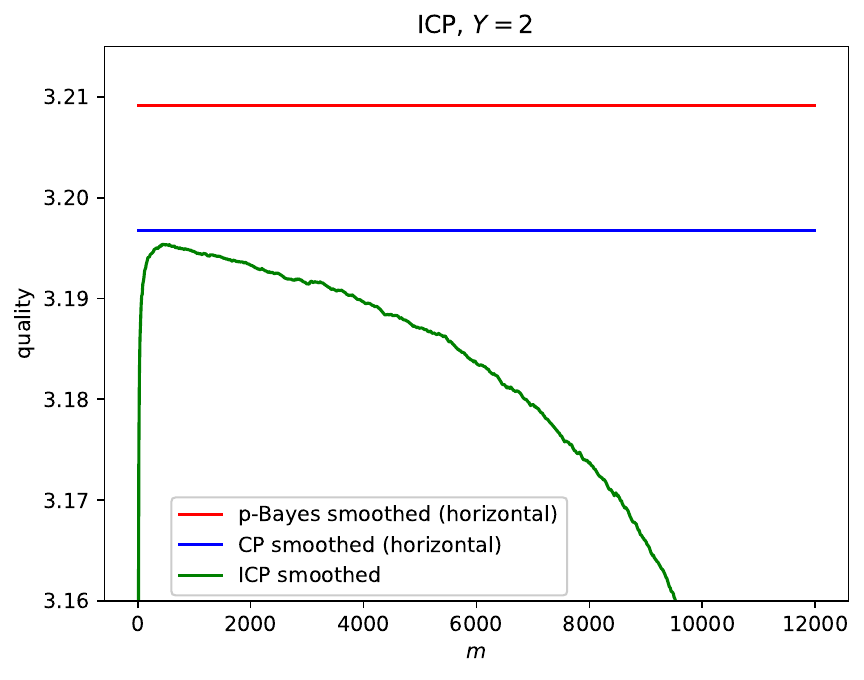}\\
    \includegraphics[width=0.48\textwidth]{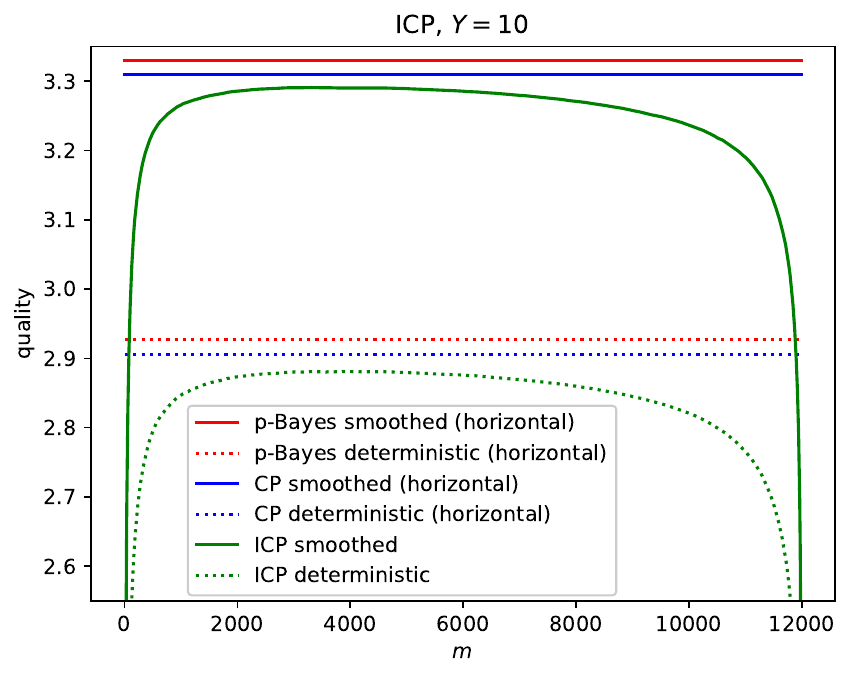}
    \includegraphics[width=0.48\textwidth]{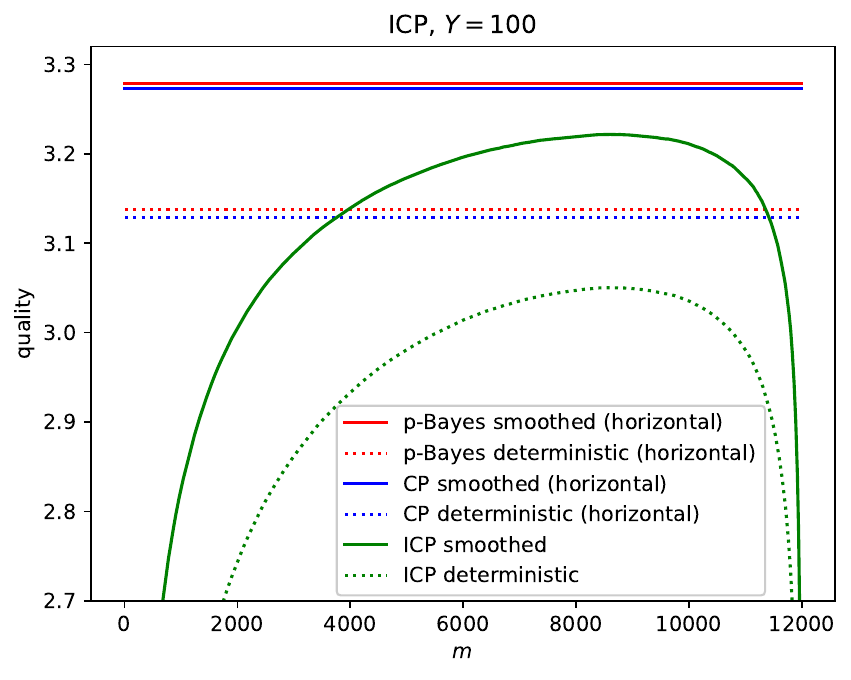}
  \end{center}
  \caption{Results for ICP; $l=12{,}000$, $Y=2$ (top), $Y=10$ (bottom left), and $Y=100$ (bottom right).
    The average value of the AFS criterion over 10,000 iterations
    is shown as function of the size $m=1,\dots,l-1$ of the proper training set.
    This is for Jeffreys's prior, $\alpha=0.5$.}
  \label{fig:ICP}
\end{figure}

Some experimental results are given in Figure~\ref{fig:ICP},
with the label ``quality'' for the vertical axis referring
to the AFS criterion with $k\to\infty$,
as mentioned earlier.
For deterministic p-values (corresponding to $\tau:=1$)
we replace the function \eqref{eq:F} by
$
  F(A,B)
  :=
  -\ln(A+B)
$.
The graphs for smoothed p-values are significantly higher than those for deterministic p-values.
The difference is especially large for $Y=2$;
in this case we do not show the graphs for deterministic p-values
since they are so far below the graphs that are shown.

Figure~\ref{fig:ICP} demonstrates that
there is no universally applicable recommendation for a good size $m'$ of the calibration set. 
For $Y=100$ the optimal fraction of calibration observations is small,
whereas for $Y=2$ it is very close to 1;
the situation for $Y=10$ is intermediate.
This is natural as learning is much quicker for $Y=2$:
in this case we have far fewer parameters to learn than for $Y=100$.

\subsection{ICEP}

In the case of inductive conformal e-predictors (ICEP),
we can again aim at optimizing either \eqref{eq:AFES} or \eqref{eq:AFES-16}.
Let us first find the ICEP corresponding to our main criterion, \eqref{eq:AFES}.
The $n'_y$ $y$s in the calibration set have
\begin{equation}\label{eq:score}
  \frac
  {1-\frac{n_y+\alpha}{m+Y\alpha}}
  {\frac{n_y+\alpha}{m+Y\alpha}}
  =
  \frac{m+Y\alpha}{n_y+\alpha} - 1
\end{equation}
as their nonconformity score (cf.\ \eqref{eq:score-test}).
Therefore, the conformal e-value for a postulated test observation $y'$ is
\begin{equation}\label{eq:e-ICP}
  e_{y'}
  =
  \frac{\frac{m+Y\alpha}{n_{y'}+\alpha} - 1}
  {
    \frac{1}{m'+1}
    \left(
      \sum_{y''\in\mathbf{Y}}
      n'_{y''}
      \left(
        \frac{m+Y\alpha}{n_{y''}+\alpha} - 1
      \right)
      +
      \left(
        \frac{m+Y\alpha}{n_{y'}+\alpha} - 1
      \right)
    \right)
  }.
\end{equation}
Overall, the value of the AFES criterion is given by \eqref{eq:overall-e},
as in the case of full conformal e-prediction.

\begin{figure}[bt]
  \begin{center}
    \includegraphics[width=0.48\textwidth]{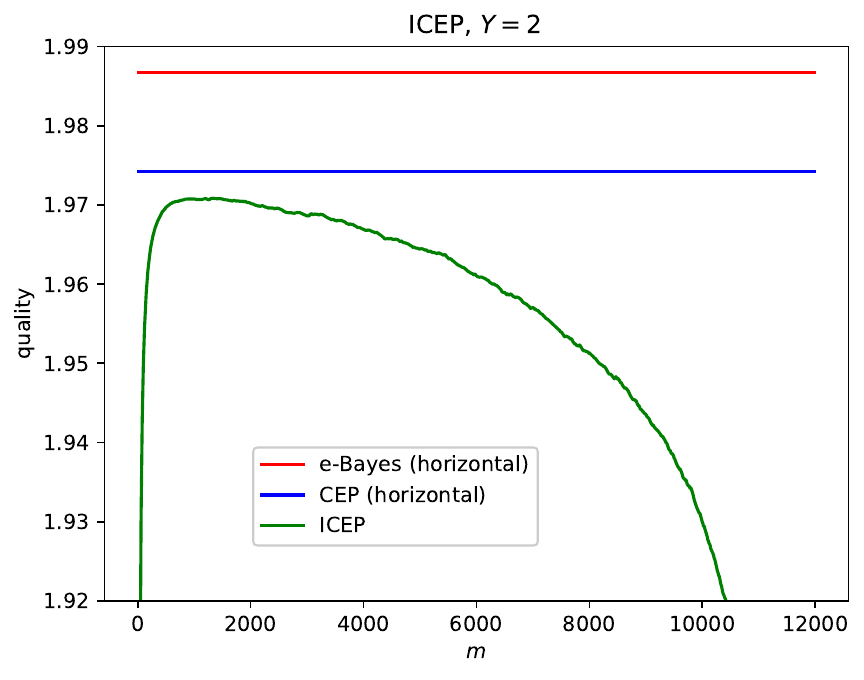}\\
    \includegraphics[width=0.48\textwidth]{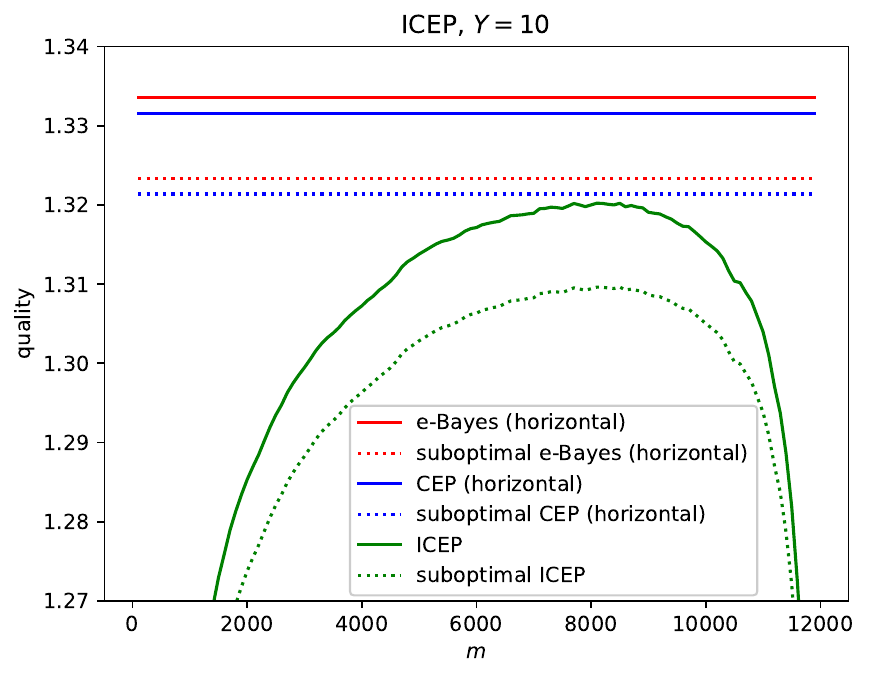}
    \includegraphics[width=0.48\textwidth]{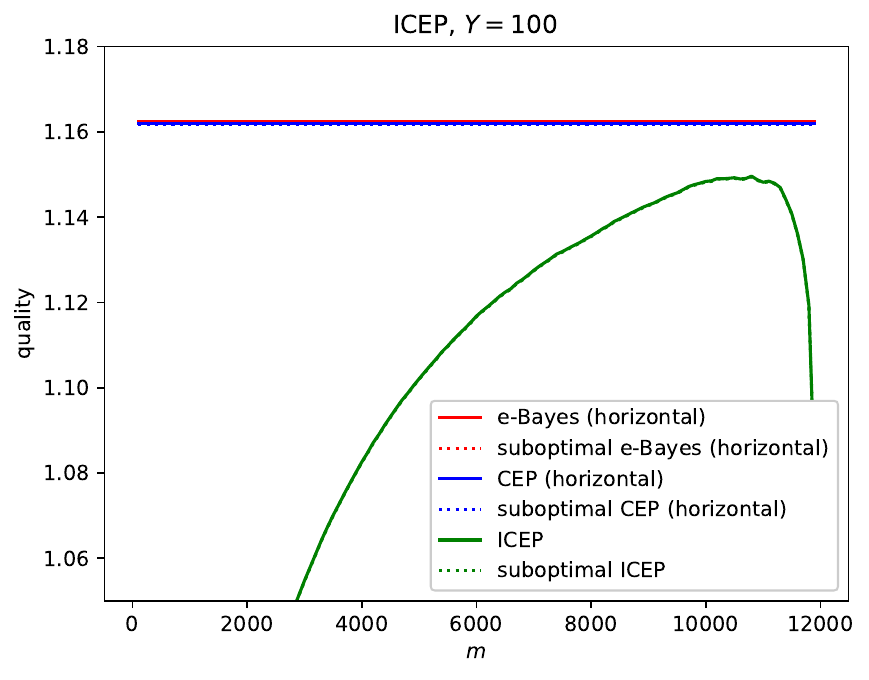}
  \end{center}
  \caption{Results for ICEP, averaged over 10,000 iterations;
    $l=12{,}000$, $Y=2$ (top), $Y=10$ (bottom left), $Y=100$ (bottom right),
    and $\alpha=0.5$.}
  \label{fig:ICEP}
\end{figure}

If we aim to optimize \eqref{eq:AFES-16},
we should drop the two subtrahends in \eqref{eq:score}
(one on the left-hand side and the other on the right-hand side)
and drop the three entries of ``${}-1$'' in \eqref{eq:e-ICP}.
We refer to this predictor as \emph{suboptimal ICEP}
since we continue to use the criterion \eqref{eq:AFES} (with $k\to\infty$)
when evaluating it in Figure~\ref{fig:ICEP}.

The analogue of Figure~\ref{fig:ICP} for e-values is given as Figure~\ref{fig:ICEP};
in the case $Y=2$ we drop the graphs for the suboptimal versions,
which are very far below the area shown.
The dotted graphs are for the suboptimal versions;
the quality of prediction as measured by \eqref{eq:overall-e}
(and shown in the plots) suffers,
but as measured by \eqref{eq:overall-e-16},
it improves (which is not shown in our plots).
The difference between the solid and dotted graphs almost disappears for $Y=100$.

\section{Cross-conformal prediction}
\label{sec:CCP}

In this section we will see that the two varieties of cross-conformal prediction,
those based on p-values and e-values,
behave very differently in our experiments.
Whereas for cross-conformal p-predictors (CCP)
the greater the number $K$ of folds the better,
as far as predictive efficiency is concerned,
for cross-conformal e-prediction (CCEP) there is a ``sweet spot'',
and moving beyond it leads to loss of predictive efficiency.
The price to pay is that the standard property of validity for full conformal p-prediction
becomes violated for CCP.

We use $K$ to denote the number of folds.
Let us assume that the size $l$ of the training set is divisible by $K$.
Then $m':=l/K$ is the ``size of the calibration set''
and $m:=l-m'$ is the ``size of the proper training set''.
For each fold $k=1,\dots,K$,
let $n'_{k,y}$ be the number of $y$s in the calibration set (i.e., fold $k$)
and $n_{k,y}$ be the number of $y$s in the proper training set
(i.e., the remaining $K-1$ folds).

The shapes of the green graphs in Figures~\ref{fig:ICP} and~\ref{fig:ICEP}
suggest that there is no universally suitable good value for $K$:
for $Y=100$ we want to include the bulk of training observations
in the proper training set (and so make $K$ large),
while for $Y=2$ we want to use as many training observations as possible for calibration
in each component inductive conformal predictor (and so set $K:=2$).
In fact this expectation is confirmed only for CCEP.

\subsection{CCP}

For each fold $k$ and each false test observation $y'$, compute
\[
  A_{k,y'}
  :=
  \sum_{y'':n_{k,y''}<n_{k,y'}} n'_{k,y''}
  \text{\qquad and\qquad}
  B_{k,y'}
  :=
  \sum_{y'':n_{k,y''}=n_{k,y'}} n'_{k,y''}.
\]
Setting
\begin{equation}\label{eq:AB}
  A_{y'} := \frac{\sum_{k=1}^K A_{k,y'}}{l+1}
  \text{\qquad and\qquad}
  B_{y'} := \frac{\sum_{k=1}^K B_{k,y'}+1}{l+1},
\end{equation}
we obtain the expression
$
  p_{y'}
  :=
  A_{y'} + \tau B_{y'}
$
for the cross-conformal p-value;
the expression for the deterministic cross-conformal p-value
is obtained by setting $\tau:=1$.
The expected p-surprisal is still \eqref{eq:expected-ps},
and the value of the AFS criterion for a given training set, its split into folds, and $\theta$
is still given by~\eqref{eq:overall-p}.

\begin{figure}[bt]
  \begin{center}
    \includegraphics[width=0.48\textwidth]{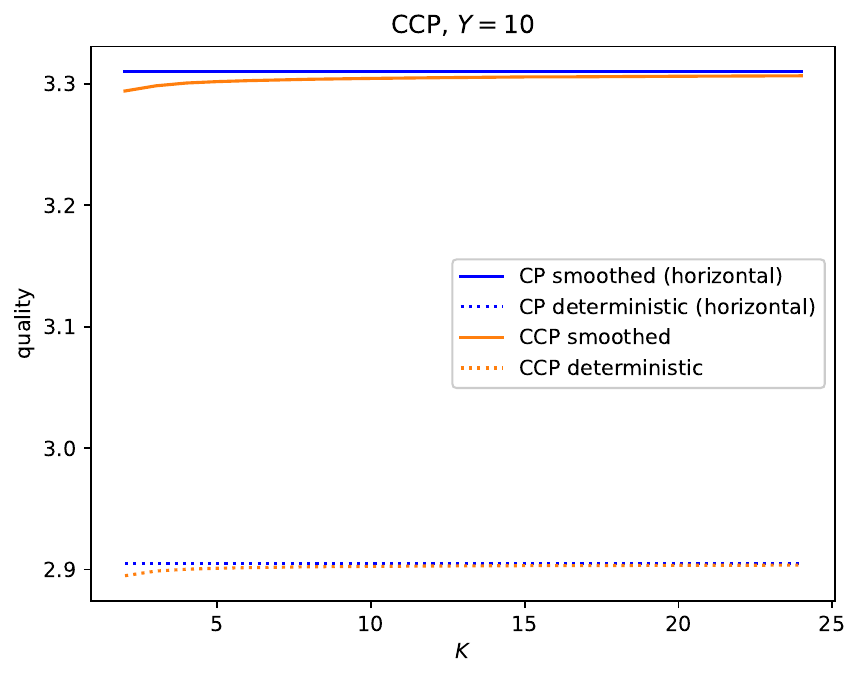}
    \includegraphics[width=0.48\textwidth]{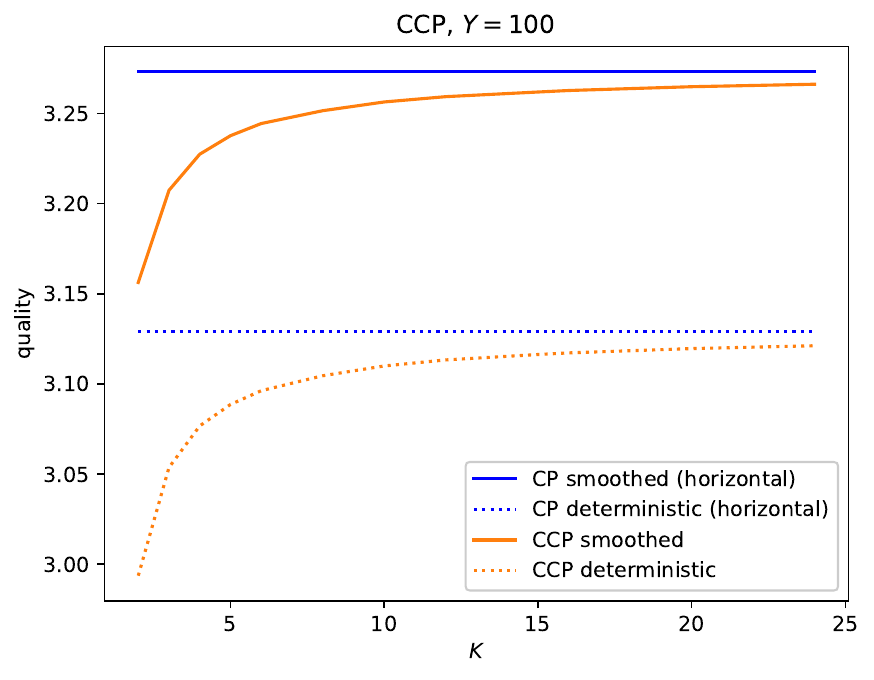}
  \end{center}
  \caption{Results for CCP; $l=12{,}000$, $Y=10$ (left), $Y=100$ (right),
    and $\alpha=0.5$.
    The value of the AFS criterion is averaged over 10,000 iterations
    and shown as function of the number of folds.}
  \label{fig:CCP}
\end{figure}

Experimental results for CCP are shown in Figure~\ref{fig:CCP}.
The AFS quality graphs for CCP are monotonic in $K$ in these experiments,
and on the right (at $K=12{,}000$, which is not shown)
they merge with the value for the full conformal predictor.
The numbers of folds in the figure are taken from the set
$K\in\{2,3,4,5,6,8,10,12,15,16,20,24\}$
(so that all $K$ are divisors of $l=12{,}000$).
The price to pay for such nice behaviour is that CCP lacks provable validity
in its usual strong form,
which may even show in experiments
(such as those involving excessive randomization;
see, e.g., \citealt[Sect.~4.8.4]{Vovk/etal:2022book}).
Let us state formally (and check in the appendix)
the coincidence of full conformal p-prediction
and CCP with the maximal number of folds (\emph{leave-one-out CCP}).

\begin{proposition}\label{prop:coincidence}
  On the same data, a leave-one-out CCP produces the same p-values
  as the corresponding full conformal p-predictor.
\end{proposition}

The case $Y=2$ is not shown in Figure~\ref{fig:CCP}
since it is even more extreme than the case of $Y=10$:
for $Y=2$ we just see two horizontal graphs.

\subsection{CCEP}

For cross-conformal e-prediction (CCEP),
we compute the e-value \eqref{eq:e-ICP}
for each fold $k$ and each false test observation $y'$,
where all $n_y:=n_{k,y}$ and $n'_y:=n'_{k,y}$ now also depend on $k$.
For each $y'$, we then average these e-values over $k=1,\dots,K$
getting one e-value $e_{y'}$.
Finally this e-value is fed into \eqref{eq:overall-e}.

\begin{figure}[bt]
  \begin{center}
    \includegraphics[width=0.48\textwidth]{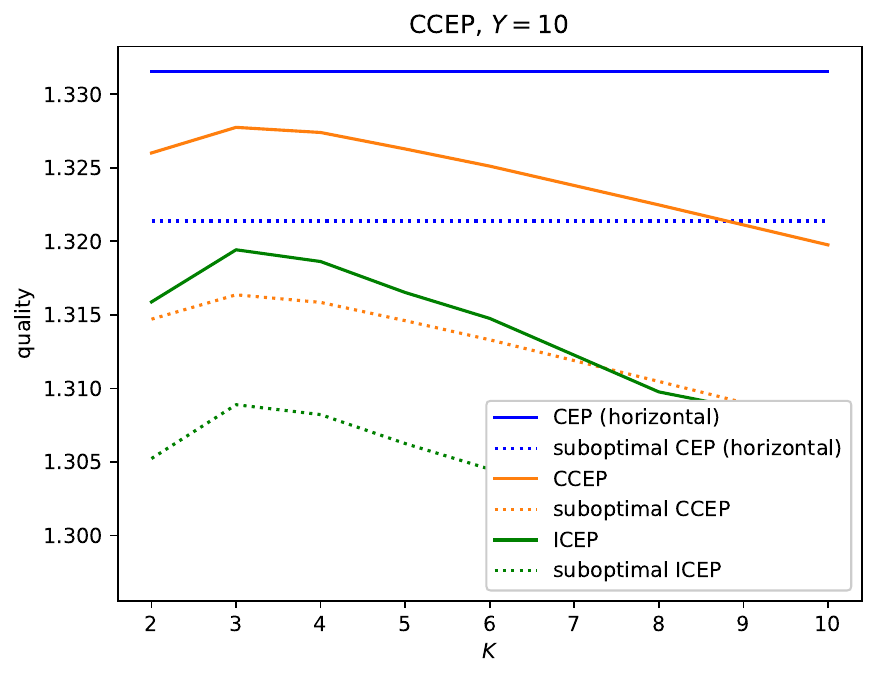}
    \includegraphics[width=0.48\textwidth]{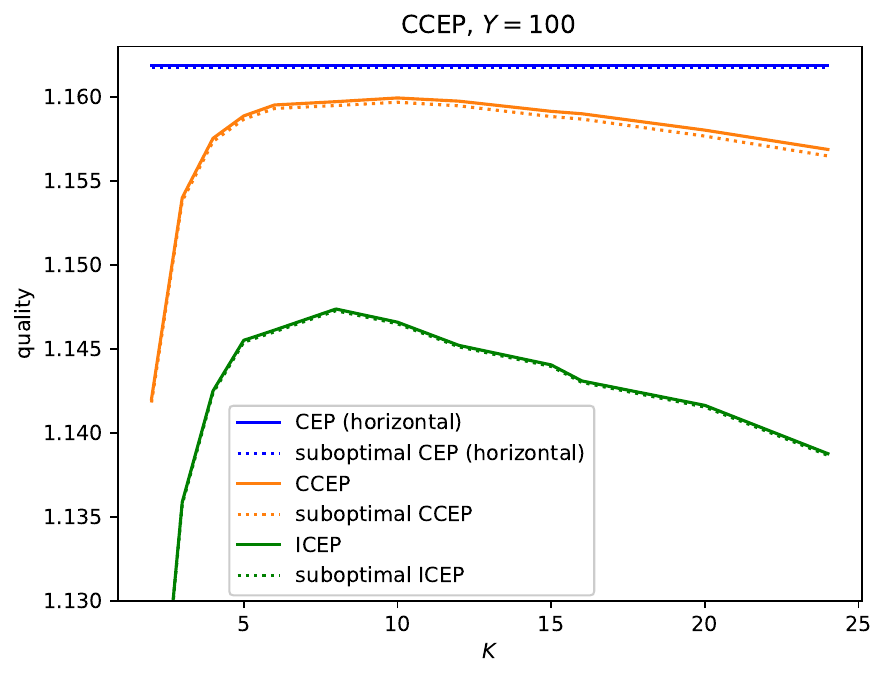}
  \end{center}
  \caption{Results for CCEP; $l=12{,}000$, $Y=10$ (left), $Y=100$ (right),
    and $\alpha=0.5$.
    The average value of the AFES criterion over 10,000 iterations
    is shown as function of the number of folds (smallest divisors of $l$).}
  \label{fig:CCEP}
\end{figure}

Experimental results for CCEP are shown in Figure~\ref{fig:CCEP}
(for $Y\in\{10,100\}$; for $Y=2$, see the next subsection).
In the left panel, $K$ ranges over the set $\{2,3,4,5,6,8,10\}$
(with the best quality attained at $K=3$),
and in the right panel, it ranges over the set $\{2,3,4,5,6,8,10,12,15,16,20,24\}$
(with the best quality attained at $K=10$).
For a smaller number $Y$ of classes,
we need fewer observations to learn a good prediction rule.
For comparison, we also show in green the AFES quality of ICEP
with the sizes $\frac{K-1}{K}l$ and $\frac{1}{K}l$ of the proper training and calibration sets,
respectively.

\subsection{Inverse CCEP}

\begin{figure}[bt]
  \begin{center}
    \includegraphics[width=0.48\textwidth]{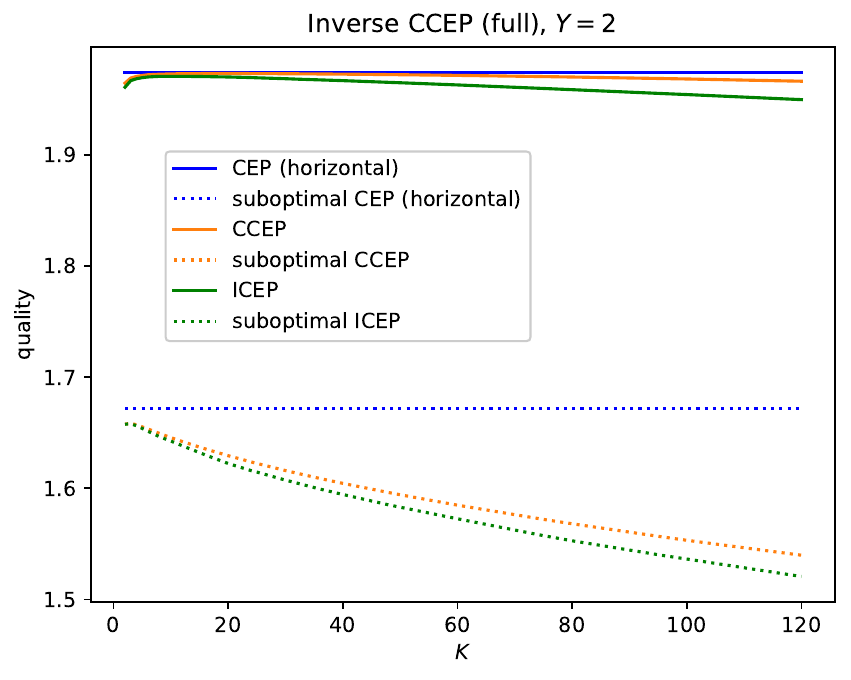}
    \includegraphics[width=0.48\textwidth]{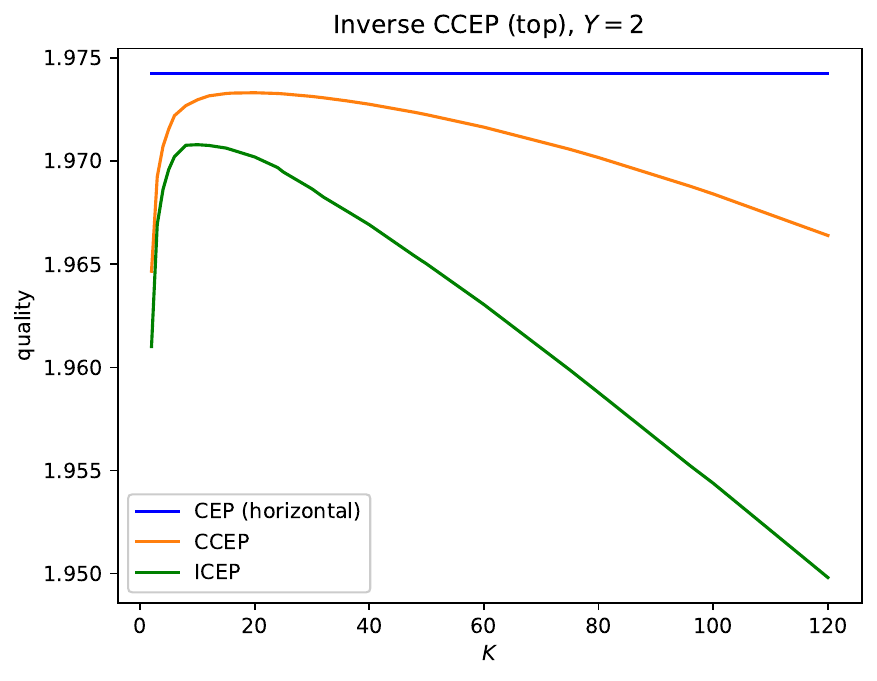}
  \end{center}
  \caption{Results for the inverse CCEP:
    the full picture is in the left panel and its top part is in the right one;
    $l=12{,}000$, $Y=2$ (binary classification), and $\alpha=0.5$.
    The average value of the AFES criterion over 10,000 iterations
    is shown as function of the number of folds (the 24 smallest divisors of $l$,
    not counting 1, here and below).}
  \label{fig:CCEPinv}
\end{figure}

For $Y=2$ (and very small $Y$ in general) the best value of $K$ is 2,
and it appears that a smaller $K$, if it were possible, would be preferable.
So in this subsection we consider the \emph{inverse CCEP}.
Each fold in turn is used as proper training set,
and the remaining folds are used for calibration;
the resulting e-values are then averaged.
See Figure~\ref{fig:CCEPinv} for experimental results;
now we use $\frac{1}{K}l$ and $\frac{K-1}{K}l$
as the sizes of proper training and calibration sets,
respectively, for ICEP
and drop ``inverse'' in the legend.
The optimal number of folds is $K=20$,
and it produces visibly better results than $K=2$,
for which CCEP and inverse CCEP coincide.

While inverse CCP would hardly ever be useful,
as we have seen above,
the situation with CCEP is very different.
Both CCEP and inverse CCEP are subsumed, to some degree,
by RICEP, which will be introduced in the next section.

\section{RICEP and BICEP}
\label{sec:BICEP}

It is clear that CCEP, even if complemented by inverse CCEP,
is too inflexible since the number $K$ of folds can only be an integer
starting from 2.
In this section we will discuss much more flexible methods.

\begin{figure}[bt]
  \begin{center}
    \includegraphics[width=0.8\textwidth]{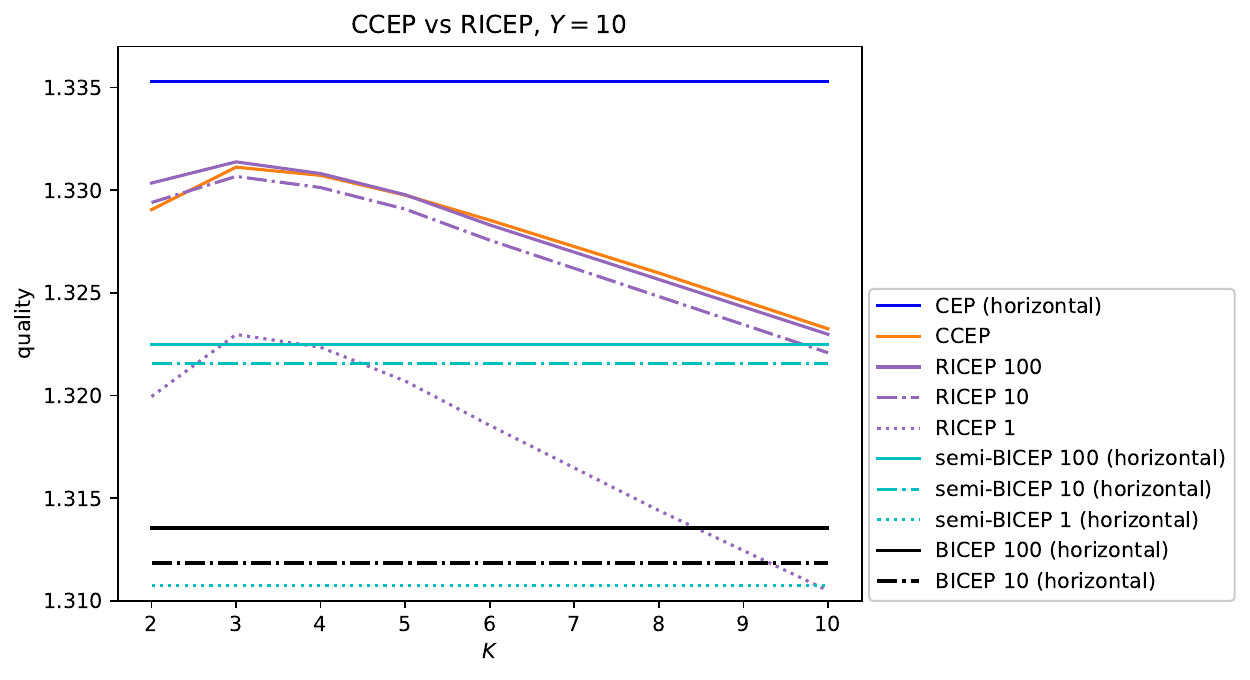}
  \end{center}
  \caption{Results for CCEP vs RICEP; $l=12{,}000$, $Y=10$, and $\alpha=0.5$.
    The average value of the AFES criterion over 10,000 iterations
    is shown as function of the number $K$ of folds (the 7 smallest divisors of $l$);
    in the case of RICEP, $m:=l-l/K$.}
  \label{fig:RICEP_Y10}
\end{figure}

\begin{figure}[bt]
  \begin{center}
    \includegraphics[width=0.8\textwidth]{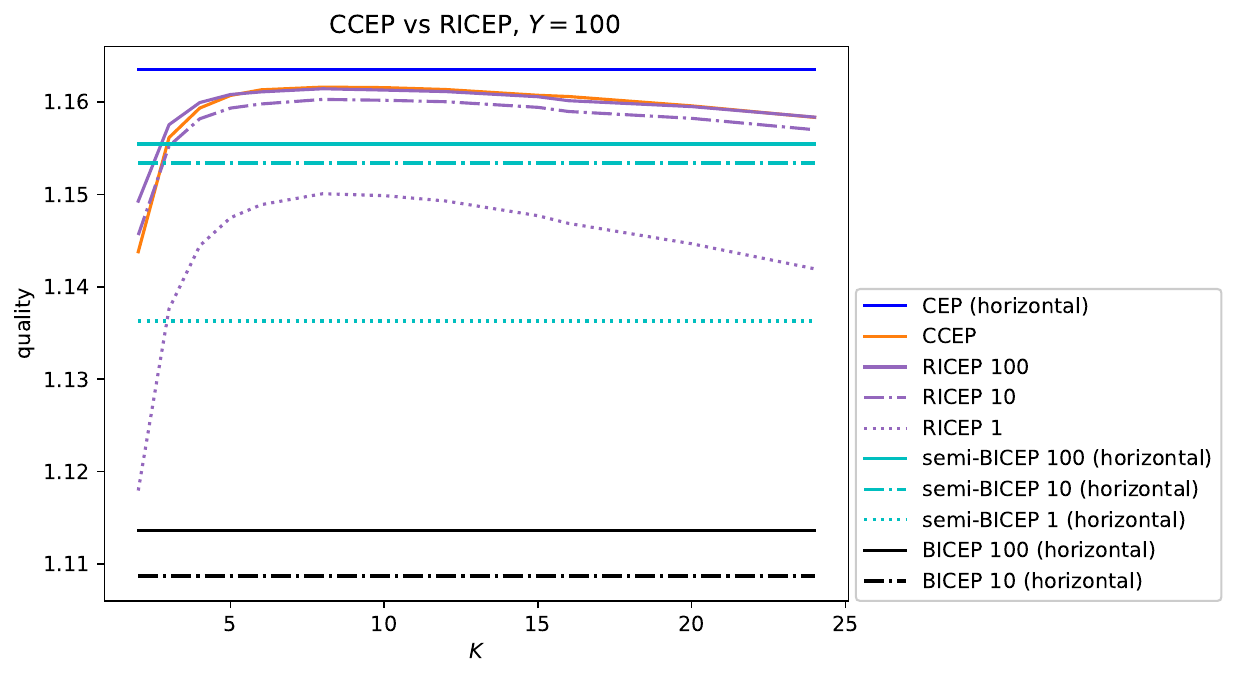}
  \end{center}
  \caption{The analogue of Figure~\ref{fig:RICEP_Y10} for $Y=100$
    and the 12 smallest divisors of $l$ as $K$.}
  \label{fig:RICEP_Y100}
\end{figure}

\begin{figure}[bt]
  \begin{center}
    \includegraphics[width=0.8\textwidth]{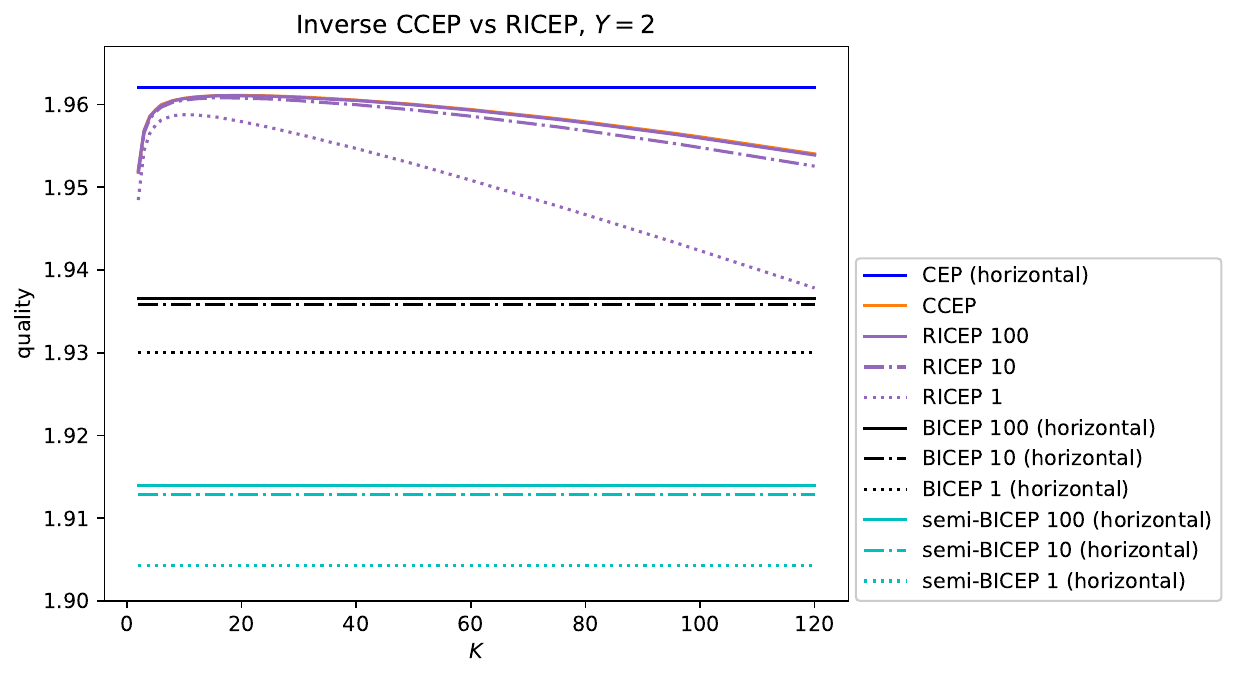}
  \end{center}
  \caption{The analogue of Figure~\ref{fig:RICEP_Y10} for $Y=2$,
    inverse CCEP,
    and the 24 smallest divisors of $l$ as $K$.}
  \label{fig:RICEP_Y2}
\end{figure}

In \emph{repeated inductive conformal e-prediction} (RICEP),
we choose the size $m\in\{1,\dots,l-1\}$ of the proper training set arbitrarily
(this is a parameter of the algorithm).
We then split the training set $N$ times randomly
into a proper training set of size $m$ and a calibration set of size $m':=l-m$,
for each split compute the corresponding e-value \eqref{eq:e-ICP},
and finally average the resulting e-values over the splits.
Therefore, there are two new parameters, $m$ and $N$.
By the law of large numbers (under mild regularity conditions),
there will not be much randomness in RICEP when $N$ is large
(unlike in ICEP);
RICEP (repeated ICEP) are ``pseudo-deterministic''.
In Figures \ref{fig:RICEP_Y10}--\ref{fig:RICEP_Y2},
we mention $N$ after ``RICEP'' and ``BICEP'' in the legend,
so that, e.g., ``RICEP 100'' stands for RICEP with $N=100$
(and we do not show BICEP 1 when it is much lower than the other graphs).

In \emph{balanced ICEP}, or \emph{BICEP},
the size of the calibration set is chosen uniformly at random from $\{1,\dots,l-1\}$,
and then the calibration set of that size is also chosen randomly;
the proper training set is then determined uniquely
as the complement of the calibration set.
In \emph{semi-BICEP},
the size of the calibration set is chosen uniformly at random from $\{1,\dots,\lfloor l/2\rfloor\}$
(so that the proper training set is at least as large as the calibration set)
and the calibration set of that size is also chosen randomly.

Figures \ref{fig:RICEP_Y10}, \ref{fig:RICEP_Y100}, and \ref{fig:RICEP_Y2}
correspond to $Y=10$, $Y=100$, and $Y=2$, respectively.
In RICEP we use $\frac{K-1}{K}l$ and $\frac{1}{K}l$
as the sizes of proper training and calibration sets,
except for Figure~\ref{fig:RICEP_Y2}, where, naturally, the sizes are swapped.
Figure~\ref{fig:RICEP_Y2} demonstrates that the full flexibility of BICEP
(as compared with semi-BICEP) is sometimes needed,
even though the performance of BICEP suffers
in Figures~\ref{fig:RICEP_Y10} and~\ref{fig:RICEP_Y100}
because of its flexibility.

In Figures \ref{fig:RICEP_Y10}--\ref{fig:RICEP_Y2},
the graphs for RICEP $N$, BICEP $N$, and semi-BICEP $N$
move up as $N$ increases,
so that the best quality is attained when $N$ is large.
This can be understood in terms of Jensen's inequality,
which implies
\begin{equation}\label{eq:Jensen}
  \ln\frac{e_1+\dots+e_N}{N}
  \ge
  \frac{\ln e_1+\dots+\ln e_N}{N}.
\end{equation}
The difference between the left-hand and the right-hand sides
is known as the \emph{Jensen gap},
and it is especially substantial when $e_1,\dots,e_N$ are very different;
to the first approximation,
the Jensen gap is proportional to the variance of $e_1,\dots,e_N$.

Inequality \eqref{eq:Jensen} explains why RICEP 10 and RICEP 100
are higher than RICEP 1.
We can also apply \eqref{eq:Jensen} to 10 in place of $N$
and to averages of 10 e-values in place of $e_1,\dots,e_{10}$,
which explains why RICEP 100 is higher than RICEP 10.
The Jensen gap between RICEP 100 and RICEP 10
is not as big as the Jensen gap between RICEP 10 and RICEP 1
since the averages of 10 e-values are less variable than the original e-values.
Of course, this argument is also applicable to BICEP and semi-BICEP.

The Jensen gap also works in favour of CCEP as compared with ICEP,
since CCEP involves averaging of the $K$ e-values coming from different folds.
(And this explains why the difference between the CCEP and ICEP graphs
tends to increase as $K$ grows in Figures~\ref{fig:CCEP} and~\ref{fig:CCEPinv}.)
Therefore, RICEP can be expected to be comparable with CCEP for $N\approx K$;
however, CCEP look slightly more efficient than that,
especially in Figure~\ref{fig:RICEP_Y10}.
In Figures~\ref{fig:RICEP_Y100} and~\ref{fig:RICEP_Y2}
the graphs for CCEP and RICEP 100 almost coincide.

\begin{figure}[bt]
  \begin{center}
    \includegraphics[width=0.6\textwidth]{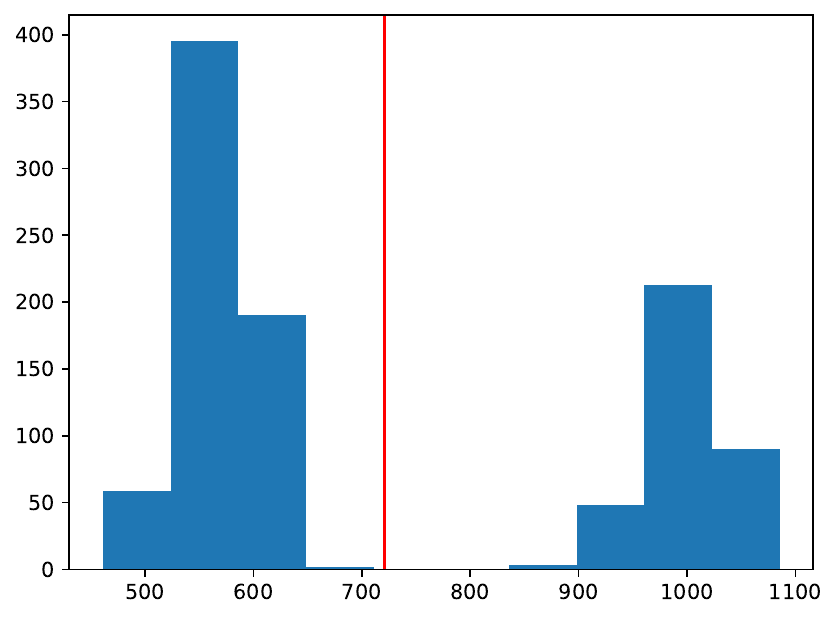}
  \end{center}
  \caption{A histogram of the e-values that RICEP averages, as described in the text.}
  \label{fig:histogram}
\end{figure}

\begin{remark}\upshape
  The Jensen gap is an artefact of our way of measuring predictive efficiency
  that involves logarithms (see \eqref{eq:AFES}).
  In principle,
  it does not mean that an improvement from, say, RICEP 1 to RICEP 100
  in Figure~\ref{fig:RICEP_Y10} is in any sense ``real''.
  (Establishing whether it is real is a direction of further research.)
  What is undoubtedly real is that the averaging in RICEP 100 produces much more stable results.
  An example of instability of e-values is shown in Figure~\ref{fig:histogram},
  which results from an experiment in which a random dataset of $12{,}000$ observations
  with $Y:=10$
  is generated from the multinomial distribution whose parameter $\theta\in\Theta$
  is generated from $\Dir_{0.5}$ (i.e., Jeffreys's prior).
  The parameter vector $\theta$ is generated only once,
  and in our experiment $\theta_3\approx7.93\cdot10^{-5}$ is its smallest component
  (and so $y=3$ is the least likely observation);
  the dataset is also generated only once.
  Then the dataset is randomly split 1000 times into proper training and calibration sets
  and the corresponding e-values \eqref{eq:e-ICP} are computed
  for the least likely observation $y=3$ (leading to high e-values).
  We take $m:=8000$ as the size of proper training sets,
  which is a reasonable value in view of the bottom left panel of Figure~\ref{fig:ICEP}.
  The histogram (in blue) shows an anomalous behaviour of the e-values;
  they are clearly separated into two clusters, above and below 800.
  The red vertical line shows the average of the 1000 e-values,
  and it is between the two clusters;
  in general, the average is still a random variable,
  but it is much more concentrated and less anomalous.
\end{remark}

\section{Conclusion}
\label{sec:conclusion}

This paper proposes new ways of using e-values in conformal prediction.
An important advantage of e-values over p-values
is that they do not require smoothing
(represented by the random number $\tau$ in \eqref{eq:p}, \eqref{eq:CP}, etc.).
Deterministic p-values perform particularly poorly in the context of this paper,
where we have no objects in a classification problem,
which leads to a large number of ties (especially for a small $Y$).
Another advantage is that CCEP have the same property of validity as CEP,
while for CCP their (provable) property of validity becomes weaker.
The advantage first described in this paper
is that CCEP can be modified to RICEP and BICEP
to address some limitations that are shared between CCEP and CCP.

These are some advantages and disadvantages of RICEP as compared with CCEP,
with ``$+$'' indicating advantages and ``$-$'' disadvantages:
\begin{itemize}
\item[$+$]
  RICEP with a large number $N$ of repetitions is pseudo-deterministic,
  while CCEP may be more volatile;
\item[$+$]
  RICEP is much more flexible;
  we do not need to choose the size of calibration sets
  from divisors (or at least approximate divisors) of the size of the training set;
  and we can easily combine different sizes, as in BICEP;
\item[$+$]
  in RICEP, the Jensen gap is under our control
  and is independent of the split proportion, unlike CCEP;
\item[$-$]
  judging by Figures~\ref{fig:RICEP_Y10}--\ref{fig:RICEP_Y2},
  CCEP may produce slightly better results for the same numbers of repetitions
  ($K$ in the case of CCEP and $N$ in the case of RICEP).
\end{itemize}

In summary,
this paper's experimental results show that RICEP has important advantages over CCEP
and should be used when it is clear \emph{a priori}
what a reasonable fraction of observations used for calibration may be.
When no or very vague prior knowledge is available
about a reasonable fraction of observations used for calibration,
BICEP is an appropriate option.
Otherwise, we could use a \emph{partial BICEP},
choosing the size of the calibration set
from some prior distribution on $\{1,\dots,l-1\}$
rather than from the uniform distribution.

\subsection*{Acknowledgments}

Many thanks to the COPA 2026 anonymous reviewers
for their helpful comments.
In the final check of the camera-ready version I used Claude Opus 5.

\appendix
\section{Some proofs}
\label{app:proofs}

\subsection{Proof of Proposition~\ref{prop:proper}}

The argument of \citet[Sect.~3.4.1, Remark 3.15]{Vovk/etal:2022book}
can be modified to cover this case as well
(it is not covered by the original argument
since the function $p\in[0,1]\mapsto-\ln p$ takes value $\infty$ at $p:=0$,
as mentioned at the end of the remark).
Namely, we can replace the second displayed equation
in \citet[Remark 3.15]{Vovk/etal:2022book} by
\begin{align}
  \sum_{y\in\mathbf{Y}}
  (-\ln p(x,y))
  &=
  \sum_{y\in\mathbf{Y}}
  \int_0^{\infty}
  1_{\{u\le-\ln p(x,y)\}}
  \dd u
  =
  \int_0^{\infty}
  \sum_{y\in\mathbf{Y}}
  1_{\{p(x,y)\le\e^{-u}\}}
  \dd u \notag\\
  &=
  \int_0^{\infty}
  \left(
    \left|\mathbf{Y}\right|
    -
    \left|
      \Gamma^{\e^{-u}}(x)
    \right|
  \right)
  \dd u
  =
  \int_0^1
  \left(
    \left|\mathbf{Y}\right|
    -
    \left|
      \Gamma^{\epsilon}(x)
    \right|
  \right)
  \frac{\dd\epsilon}{\epsilon}.
  \label{eq:3.15}
\end{align}
The last integral converges since
$
  \left|\mathbf{Y}\right|
  -
  \left|
    \Gamma^{\epsilon}(x)
  \right|
  =
  0
$
for sufficiently small $\epsilon>0$.

\subsection{Proof of Proposition~\ref{prop:Bayes-p-optimality}}

Proposition~\ref{prop:Bayes-p-optimality} can be easily deduced
from \citet[Theorem~3.1]{Vovk/etal:2022book}
and the modification \eqref{eq:3.15} of the argument in 
\citet[Remark~3.15]{Vovk/etal:2022book}.
Under our Bayesian model and conditionally on the training set,
we are in the idealized setting of \citet[Sect.~3.1.4]{Vovk/etal:2022book}
with the data-generating distribution equal to the predictive distribution
for the test observation.
According to \citet[Theorem~3.1]{Vovk/etal:2022book},
the probability measure \eqref{eq:P} is an N-optimal idealized conformity measure
conditionally on the training set.
This implies that it is N-optimal unconditionally,
which in combination with \eqref{eq:3.15} implies that the p-values \eqref{eq:p}
solve \eqref{eq:Bayes-p-optimality}.

In general, a p-variable solving \eqref{eq:Bayes-p-optimality} is not unique
since, according to \citet[Theorem~3.1]{Vovk/etal:2022book},
we can replace the probability measure \eqref{eq:P}
used in \eqref{eq:p} in the role of idealized conformity measure
by its refinement;
this will lead to the same value of \eqref{eq:Bayes-p-optimality}.

\subsection{Proofs of Propositions~\ref{prop:Bayes-e-optimality} and~\ref{prop:Bayes-e-optimality-16}}

To prove Proposition~\ref{prop:Bayes-e-optimality},
apply Proposition~15 in \citet{Vovk:2025PR} conditionally on the training set.
Then \eqref{eq:Bayes-e-optimality} holds conditionally on the training set
and, therefore, unconditionally.

To show that there is a unique solution to \eqref{eq:Bayes-e-optimality},
suppose there are two different solutions.
Then their arithmetic mean will provide
a better value for the objective function in \eqref{eq:Bayes-e-optimality}
by Jensen's inequality.
The value will be strictly better 
since every finite sequence of observations has a positive probability
under our assumptions.

The proof of Proposition~\ref{prop:Bayes-e-optimality-16} is a simple modification;
cf.\ \citet[Remark~16]{Vovk:2025PR}.

\subsection{Proof of Proposition~\ref{prop:coincidence}}

We need to check that the $A_{y'}$ and $B_{y'}$ defined by \eqref{eq:CP}
coincide with those defined by \eqref{eq:AB} when $K=l$.
Let us first do it for the two $A_{y'}$.
Since $n'_{k,y''}$ is the indicator function of $y''$ being the observation in the $k$th fold,
$A_{k,y'}$ is the indicator function of $n_{k,y_k}<n_{k,y'}$.
Since the last inequality is impossible for $y_k=y'$,
$A_{k,y'}$ is the indicator function of the conjunction of $y_k\ne y'$ and $n_{y_k}-1<n_{y'}$.
Therefore, the sum of $A_{k,y'}$ over $k$ coincides with the numerator of $A_{y'}$ in \eqref{eq:CP}.

Now let us do it for the two $B_{y'}$.
Notice that $B_{k,y'}$ is the indicator function of $n_{k,y_k}=n_{k,y'}$,
this equality is true for $y_k=y'$,
and this equality is equivalent to $n_{y_k}-1=n_{y'}$ when $y_k\ne y'$.
Therefore, the sum of $B_{k,y'}$ over $k$ coincides
with the numerator of $B_{y'}$ in \eqref{eq:CP} apart from the ``${}+1$''.
\end{document}